\documentclass[conference]{IEEEtran}
\IEEEoverridecommandlockouts
\usepackage{cite}
\usepackage{amsmath,amssymb,amsfonts}
\usepackage{algorithmic}
\usepackage{graphicx}
\usepackage{textcomp}
\usepackage{url}
\usepackage{xcolor}
\usepackage{caption}      
\usepackage{subcaption}
\usepackage{fancyhdr}
\usepackage{arabtex}
\usepackage{utf8}

\def\BibTeX{{\rm B\kern-.05em{\sc i\kern-.025em b}\kern-.08em
    T\kern-.1667em\lower.7ex\hbox{E}\kern-.125emX}}
\makeatletter
\let\old@ps@headings\ps@headings
\let\old@ps@IEEEtitlepagestyle\ps@IEEEtitlepagestyle
\def\confheader#1{%
\def\ps@IEEEtitlepagestyle{
\old@ps@IEEEtitlepagestyle
\def\@oddhead{\strut\hfill#1\hfill\strut}
\def\@evenhead{\strut\hfill#1\hfill\strut}
}
\ps@headings
}
\makeatother
\confheader{
\small{Proceedings of the 12th RSI International Conference on Robotics and Mechatronics (ICRoM 2024), Dec. 17-19, 2024, Tehran, Iran} 
}
\usepackage[pscoord]{eso-pic}
\newcommand{\placetextbox}[3]{
\setbox0=\hbox{#3}
\AddToShipoutPictureFG*{ \put(\LenToUnit{#1\paperwidth},\LenToUnit{#2\paperheight}){\vtop{{\null}\makebox[0pt][c]{#3}}}
}
}
\placetextbox{.23}{0.055}{\textbf{\small{979-8-3315-2973-4/24/\$31.00~\copyright 2024 IEEE}}}
  
\begin{document}
\title{Integrating Persian Lip Reading in Surena-V Humanoid Robot for Human-Robot Interaction
}

\author{\IEEEauthorblockN{Ali Farshian Abbasi}
\IEEEauthorblockA{\textit{Center of Advanced Systems} \\ \textit{and Technologies (CAST)} \\
\textit{School of Mechanical Engineering} \\
\textit{University of Tehran}\\
alifarshian@ut.ac.ir}
\and
\IEEEauthorblockN{Aghil Yousefi-Koma}
\IEEEauthorblockA{\textit{Center of Advanced Systems} \\ \textit{and Technologies (CAST)} \\
\textit{School of Mechanical Engineering} \\
\textit{University of Tehran}\\
aykoma@ut.ac.ir}
\and
\IEEEauthorblockN{Soheil Dehghani Firouzabadi}
\IEEEauthorblockA{\textit{Center of Advanced Systems} \\ \textit{and Technologies (CAST)} \\
\textit{School of Mechanical Engineering} \\
\textit{University of Tehran}\\
soheil.dehghani@ut.ac.ir}
\and
\IEEEauthorblockN{Parisa Rashidi}
\hspace{0.52\textwidth}
\IEEEauthorblockA{\textit{Center of Advanced Systems} \\ \textit{and Technologies (CAST)} \\
\textit{School of Mechanical Engineering} \\
\textit{University of Tehran}\\
Parisa.rashidi@ut.ac.ir}
\and
\IEEEauthorblockN{Alireza Naeini}
\IEEEauthorblockA{\textit{Center of Advanced Systems} \\ \textit{and Technologies (CAST)} \\
\textit{School of Mechanical Engineering} \\
\textit{University of Tehran}\\
alirezaa.naaeini@ut.ac.ir}
\and
\IEEEauthorblockN{                                         }
\IEEEauthorblockA{\textit{                                   } \\
\textit{                       }\\
             \\
                    }
}

\maketitle

\begin{abstract}
Lip reading is vital for robots in social settings, improving their ability to understand human communication. This skill allows them to communicate more easily in crowded environments, especially in caregiving and customer service roles.
Generating a Persian Lip-reading dataset, this study integrates Persian lip-reading technology into the Surena-V humanoid robot to improve its speech recognition capabilities. Two complementary methods are explored, an indirect method using facial landmark tracking and a direct method leveraging convolutional neural networks (CNNs) and long short-term memory (LSTM) networks. The indirect method focuses on tracking key facial landmarks, especially around the lips, to infer movements, while the direct method processes raw video data for action and speech recognition. The best-performing model, LSTM, achieved 89\% accuracy and has been successfully implemented into the Surena-V robot for real-time human-robot interaction. The study highlights the effectiveness of these methods, particularly in environments where verbal communication is limited.
\end{abstract}


\section{Introduction}
Lip reading, or speech reading, is the skill of understanding spoken language by observing lip, facial, and tongue movements without hearing the words. This complex ability is especially useful in noisy settings or for those with hearing impairments [1]. Lip reading plays a critical role in both human-to-human and human-computer interactions, acting as an essential communication tool for individuals with verbal limitations [2],[3].

Without auditory cues, visual signals become crucial for recognizing speech. However, lip reading has inherent challenges, such as differentiating between homophones [4] and handling variations in lip movements for the same phrases among different speakers. Additionally, the orientation of a person’s face can significantly affect recognition [5]. Humanoid robots, created to replicate human form and behaviors, are increasingly prevalent across various industries. These robots are designed to take on tasks traditionally performed by humans [6]. With ongoing technological advancements, humanoid robots are proving valuable in settings where human-like interaction is essential. By integrating lip-reading recognition technology, these robots can interpret visual cues when verbal communication is obstructed, such as at busy events or in noisy industrial spaces. This function supports smooth interaction and enhances efficiency in crowded environments [7].
\begin{figure*}[hbtp]  
    \centering
    \begin{minipage}{0.44\textwidth}
        \centering
        \includegraphics[width=\linewidth]{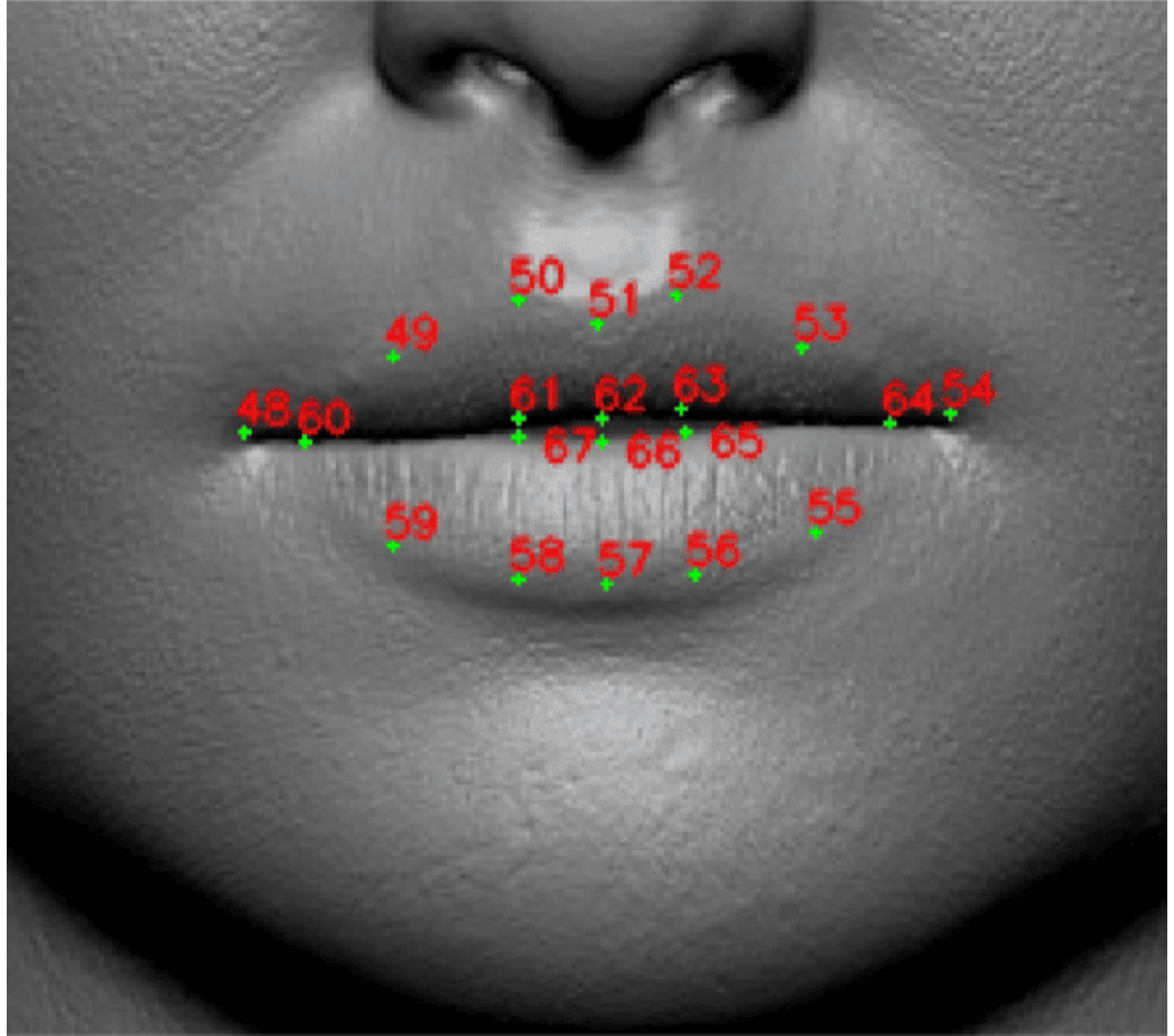}  
    \end{minipage}
    \hspace{0\textwidth}  
    \begin{minipage}{0.44\textwidth}
        \centering
        \includegraphics[width=\linewidth]{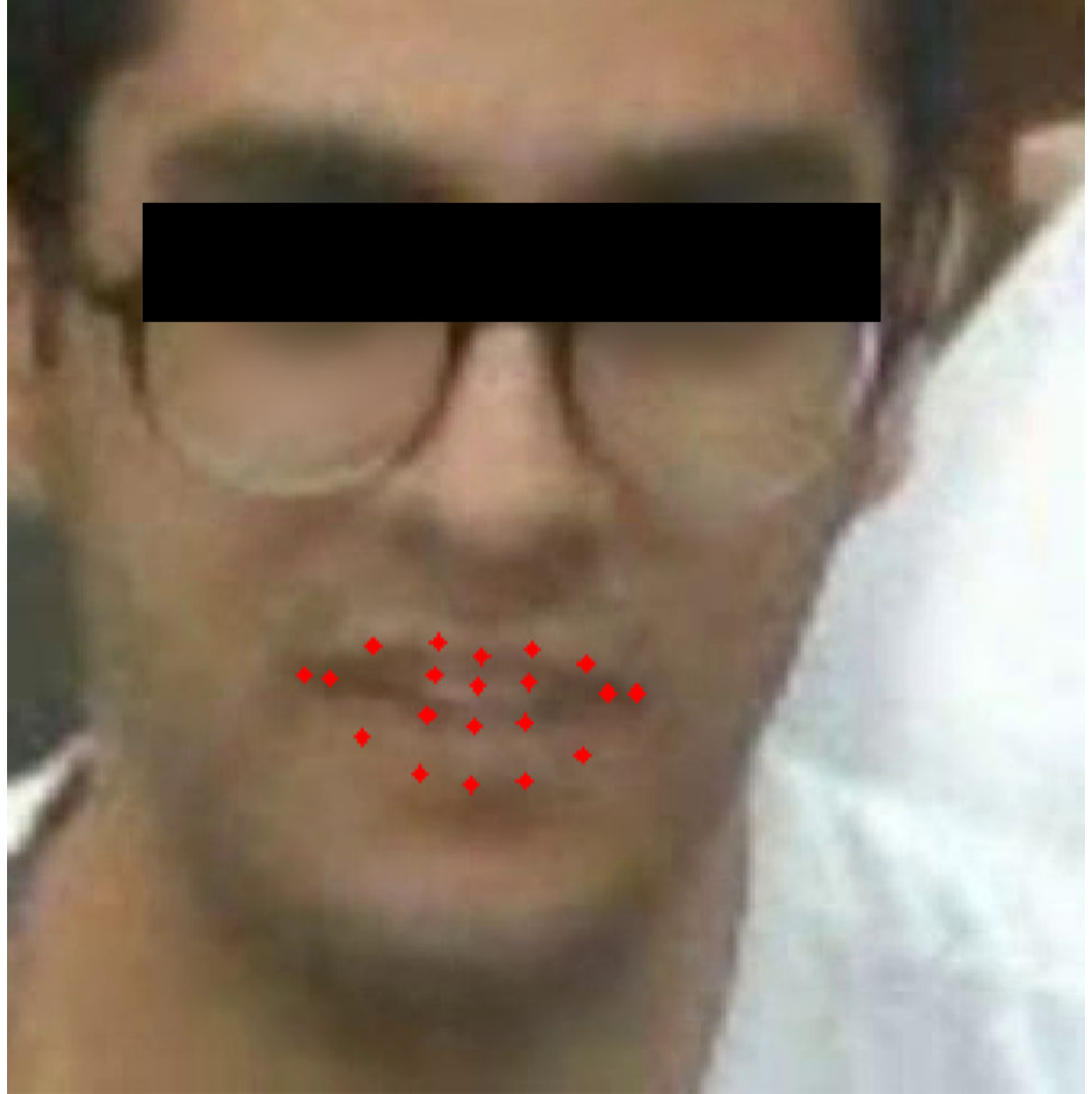}  
    \end{minipage}
    
    \vspace{0.008\textwidth}  
       
    \caption{The location of marks and their index in real and 3D model picture}
\end{figure*}

Considerable research has been dedicated to advancing lip-reading technology, leading to various datasets and methodologies. For example, Javad Peymanfard et al. [8] created a Persian word dataset of 244,000 videos featuring celebrities from the Persian platform Aparat. This study used the AV-Hubert model, achieving 21.43\% accuracy. Similarly, Mahsa Hedayatipour et al. [9] developed a Persian dataset with 30 hours of video from 40 speakers, using a Hidden Markov Model (HMM) to classify six Persian words with 47.73\% accuracy.
Sivri created a Turkish dataset, containing three phrases and three words sourced from Turkish YouTube videos [10]. The dataset was evaluated using Convolutional Neural Networks (CNN), Long Short-Term Memory Networks (LSTM), and Bidirectional Gated Recurrent Units (BGRU). Aripin et al. [11] compiled an Indonesian lip-reading dataset containing ten words and four phrases spoken by ten participants, reaching 95\% accuracy with Long Short-Term Memory Networks (LRCN). Faisal et al. [12] built an Urdu dataset featuring ten participants who repeated ten words and phrases ten times each. Although phrase recognition accuracy was challenging, the LSTM model achieved 62\%. Dweik produced an Arabic dataset of 1,051 videos with male and female speakers enunciating common words under various lighting and angle conditions, achieving up to 82.86\% accuracy using three different deep-learning models [13].

This study gathered a comprehensive dataset at the Centre for Advanced Science and Technology (CAST), recorded by laboratory research students. Various classification techniques were tested for their ability to recognize lip movements, identifying the most accurate method, which was subsequently applied to the Surena V humanoid robot. This technology allows the robot to recognize words and respond accordingly with gestures, facial expressions, or vocalizations that match the intended message. This work demonstrates the potential of advanced lip-reading technology to improve human-robot interactions in environments where verbal communication is limited.

The following sections introduce the dataset, explain the methodology, and detail the approaches tested. The next step involves implementing the best method on the Surena V robot. Finally, the study’s results and their implications for human-robot interaction are discussed.

\section*{Dataset}
 
For Persian, the lipreading model has been trained to recognize seven common words, as shown in Table I, along with their English pronunciations and meanings. These words were selected based on criteria such as usage frequency, simplicity, coherence, and their applicability for commanding the Surena humanoid robot. Data for this project was collected using the robot’s frontal RGB-D camera, recording videos at 20 fps to enhance model accuracy.
An RGB-D camera captures both color RGB and depth D information in real-time, combining traditional 2D image data with depth perception using infrared sensors. This enables it to create 3D representations of objects and environments, making it ideal for applications in robotics, AR/VR, and 3D scanning[14]

The study involved 20 participants (16 men and 4 women), each pronouncing the selected words 20 times with a 1-second interval between repetitions. The original videos were segmented so that each clip contained a single pronounced word, resulting in 20 smaller videos per recording. Initially, the videos had a resolution of 1280×720, which was suboptimal for tracking lip movement. To improve focus on the lips, the Cvzone face detector was employed, reducing the video size to 300×300. After these adjustments, the total number of video clips increased to 2,800. Unlike other datasets, this dataset emphasizes facial features during speech, acknowledging that lipreading accuracy can improve by focusing on the entire face, as speaking involves more than just the lips.also Table II provides a summary of key aspects from datasets mentioned in introduction.

\begin{table}[b]
\caption{Wordset of persian dataset}
\begin{center}
\begin{tabular}{|l|l|l|}
\hline
English Pronunciation & English meaning \\
\hline
Salam & Hello \\
\hline
Bro & Go \\
\hline
Bia & Come \\
\hline
Khodahafez & Goodbye \\
\hline
Surena & Surena \\
\hline
Begir & Take \\
\hline
Benevis & Write \\
\hline
\end{tabular}
\label{tab1}
\end{center}
\end{table}
\section*{Methodology}

\begin{table}[b]
\caption{Comparison of datasets}
\begin{center}
\begin{tabular}{|c|c|c|c|c|}
\hline
Language & Speakers & Classes & Total data & resolution \\ \hline
Urdu     & 10       & 18      & 2000       & $100\times50$    \\ \hline
Arabic   & 73       & 10      & 1051       & $66\times100$    \\ \hline
Indonesian & 8      & 14      & 4000       & $640\times480$   \\ \hline
Persian(ours)  & 20       & 7       & 2800       & $300\times300$   \\ \hline
Persian & 52         & 10     & 1560       & $26\times44$ \\ \hline
\end{tabular}
\label{tab1}
\end{center}
\end{table}
This study explores two distinct approaches for lip-reading detection: an indirect method and a direct method. The indirect method relies on the tracking of facial landmarks to infer lip movements, while the direct method is anticipated to provide a more straightforward approach to data extraction and processing. The specifics of the direct method are still being finalized, but both approaches are crucial for understanding the overall efficacy of the lip-reading detection system.\

\begin{figure*}[hbtp]  
    \centering
    \begin{minipage}{0.44\textwidth}
        \centering
        \includegraphics[width=\linewidth]{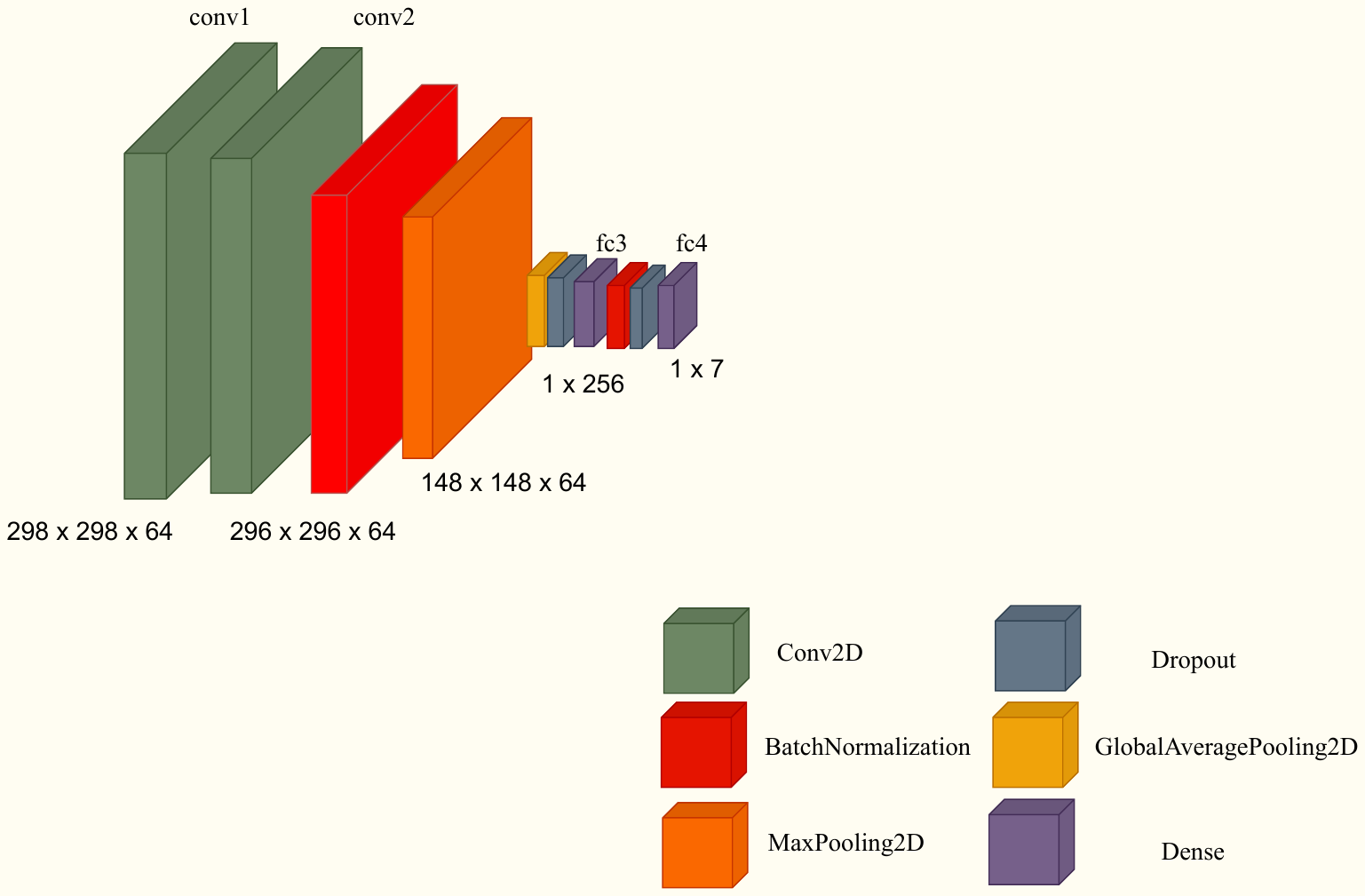}  
        \subcaption{}  
    \end{minipage}
    \hspace{0.008\textwidth}  
    \begin{minipage}{0.44\textwidth}
        \centering
        \includegraphics[width=\linewidth]{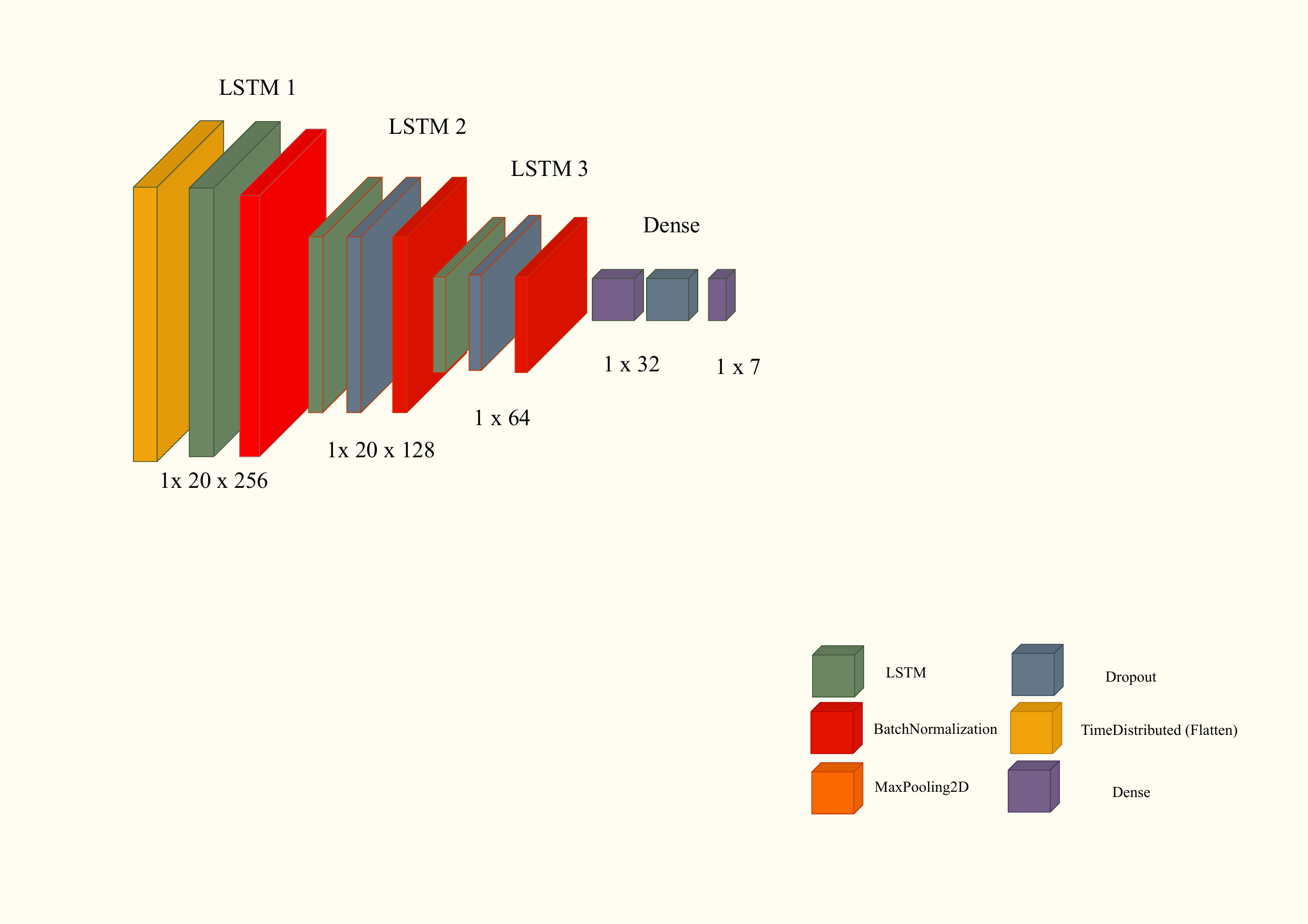}  
        \subcaption{}  
    \end{minipage}
    
    \vspace{0.02\textwidth}  
       
    \caption{The structure of the (a) CNN and (b) LSTM Neural Networks in direct method}
\end{figure*}

\noindent I. Indirect Method

In the indirect approach, lip-reading detection is based on the extraction and analysis of key facial landmarks over time. The dlib library is employed to identify and locate facial landmarks, specifically focusing on points 48 to 68, which correspond to the contour and shape of the lips. These landmarks are meticulously tracked across a series of video frames to capture the intricate movements of the lips during speech which can be seen in Fig. 1.
For each frame in the video, the spatial coordinates of the identified lip points are extracted and organized into a 3D array. This array is structured as (x, y, frame), where the x and y dimensions represent the spatial location of each lip point, and the frame dimension indicates the sequence of time. Given that the videos in the dataset vary significantly in length, a standardized portion of each video, specifically one second or 20 frames, is selected to ensure uniformity in input size across the dataset. This standardization facilitates consistent processing by the neural network .
Once the 3D array has been generated, it is fed into a selection of convolutional neural network (CNN) models for classification. Several well-established CNN architectures are tested, including VGG19 and ResNet, which are known to perform adequately for similar tasks.[15]
Before the arrays are processed by the models, a series of preprocessing steps are applied to enhance the robustness of the system. Normalization techniques are employed to mitigate the effects of potential noise factors, such as variations in lighting conditions, slight head movements, and changes in camera angles. By minimizing these extraneous influences, the model can focus on the relevant features of lip movement, thereby improving overall accuracy.
Following preprocessing, the MobileNet model is trained to recognize patterns in lip movements that correspond to specific phonetic units or words. The training process is conducted using the Adam optimizer and a cross-entropy loss function, with multiple epochs employed to ensure convergence and optimal performance. The system’s efficacy is evaluated by comparing the predicted text output with the actual spoken words, and it is consistently found that MobileNet achieves the highest levels of accuracy and speed among the tested architectures.

\begin{figure*}[hbtp]  
    \centering
    
    \begin{minipage}{0.45\textwidth}
        \centering
     \includegraphics[width=0.90\linewidth,height=0.24\textheight]{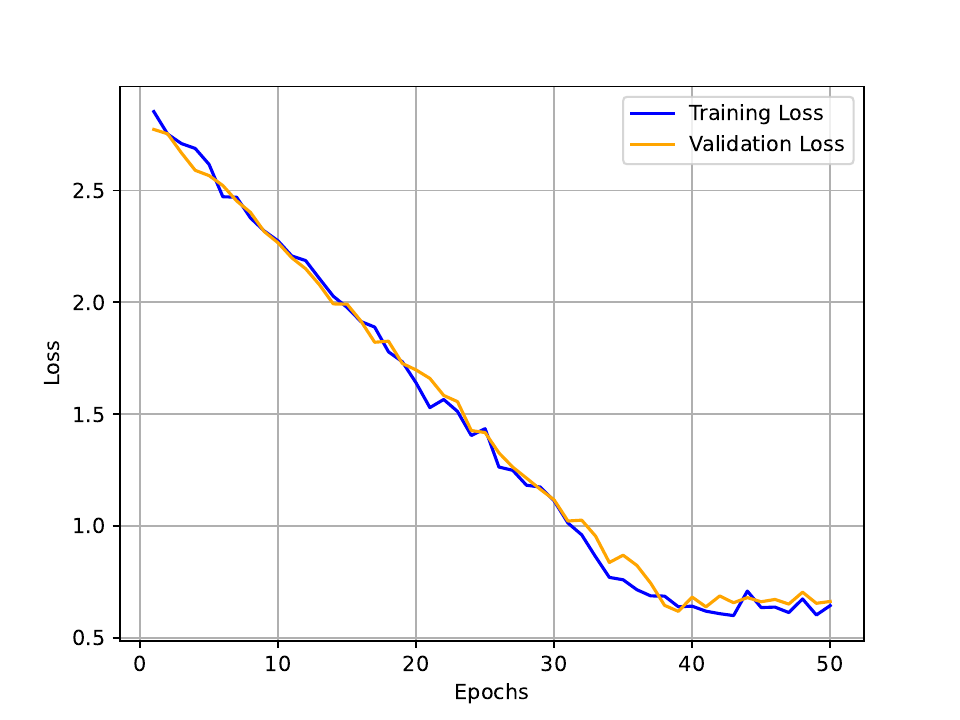}
    \end{minipage}
    \hspace{0\textwidth}  
    \begin{minipage}{0.45\textwidth}
        \centering      \includegraphics[width=0.90\linewidth,height=0.24\textheight]{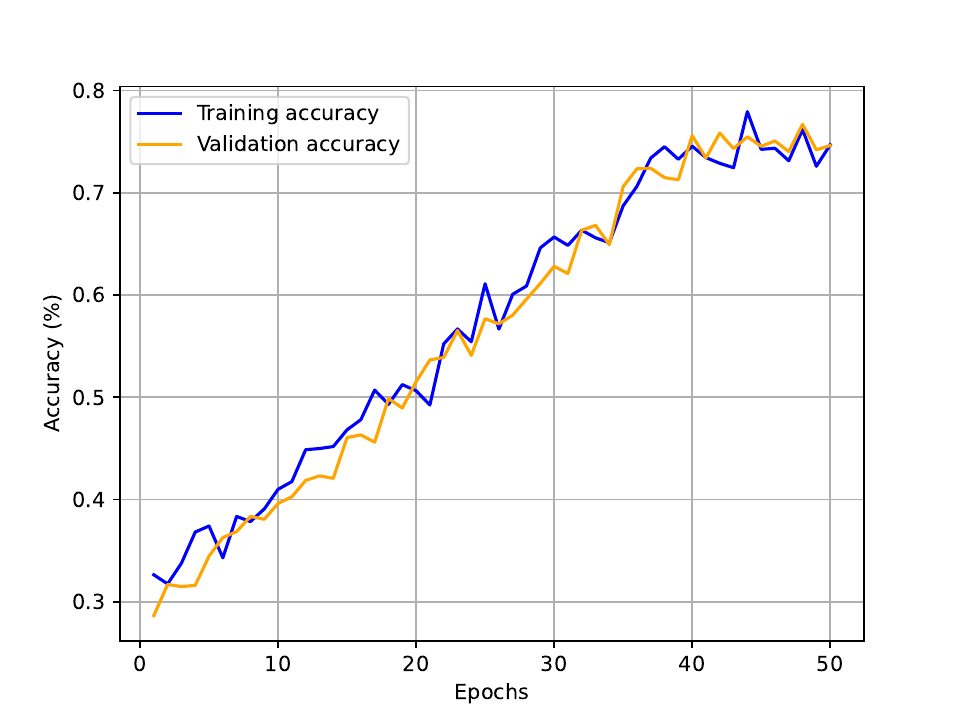}  
    \end{minipage}
    
    \vspace{-0.2cm}  
    
    \subsection*{(a)}
    
    \vspace{0.005\textwidth}  
    
    
    \hspace{0.026\textwidth}
    \begin{minipage}{0.46\textwidth}

    \includegraphics[width=0.90\linewidth,height=0.22\textheight]{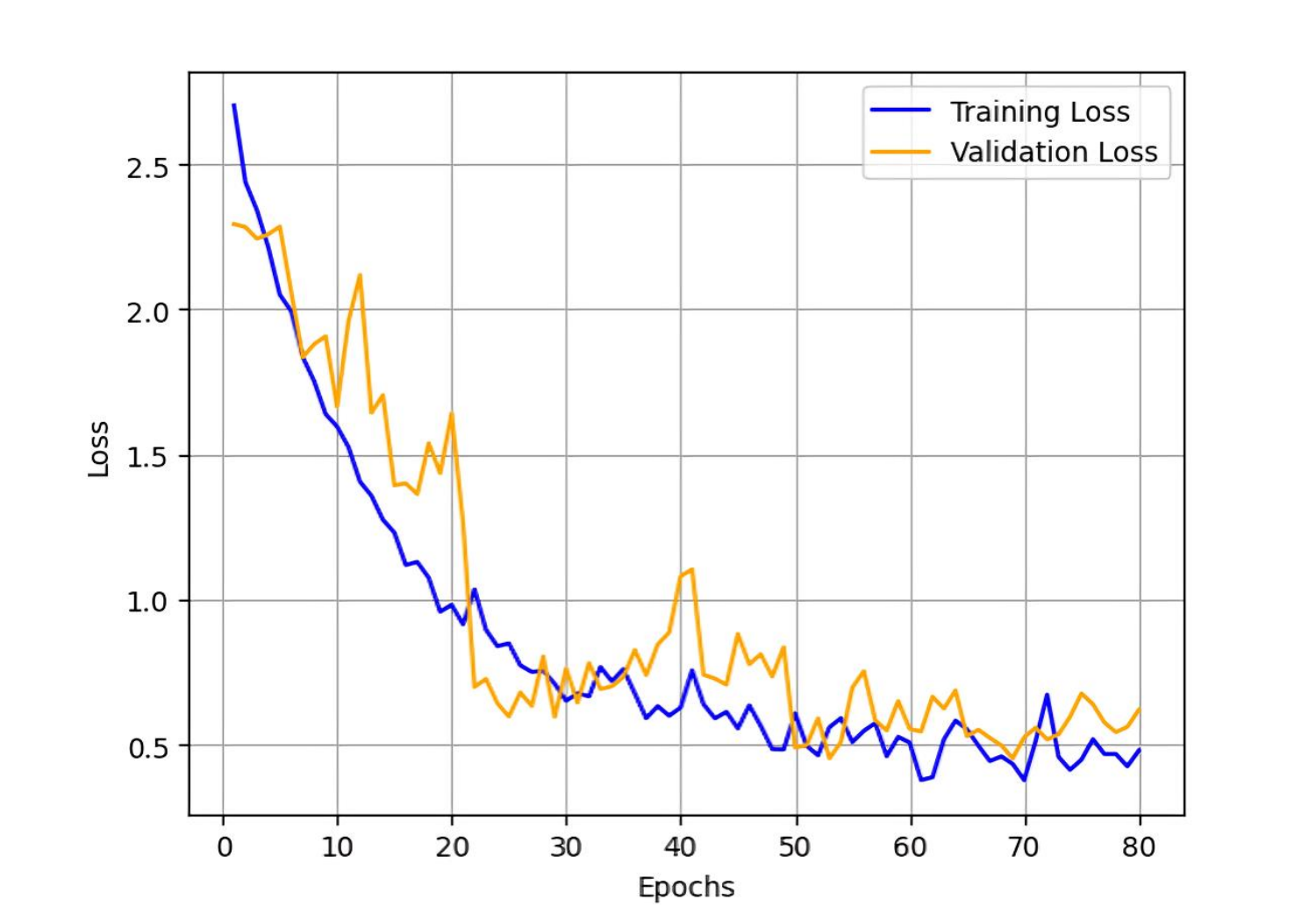}

    \end{minipage}
    \hspace{-0.014\textwidth}
    \begin{minipage}{0.46\textwidth}

    \includegraphics[width=0.92\linewidth,height=0.23\textheight]{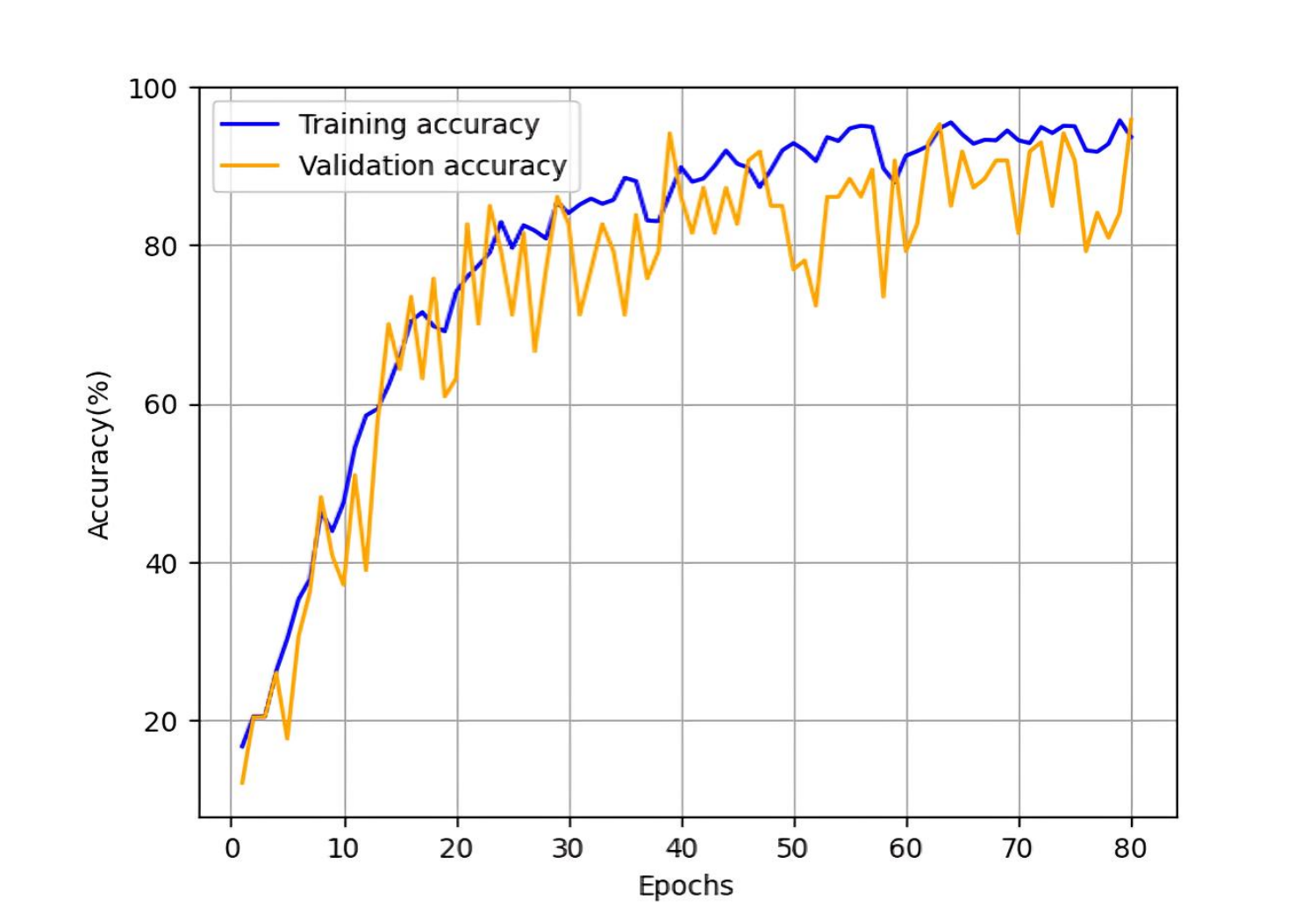}
    \end{minipage}
    
    \vspace{-0.2cm}  
    
    \subsection*{(b)}
    
    \caption{The accuracy and loss for (a) CNN  and (b) LSTM approach in direct method}
\end{figure*}

\vspace{0.1cm}

\noindent II.Direct Method

Two distinct direct approaches Convolutional Neural Networks (CNN) and Long Short-Term Memory (LSTM) networks have been implemented for action classification and verbal command recognition.

\subsection{CNN Approach}

In the first approach, we developed and evaluated a CNN for classifying actions in videos. The dataset consisted of action videos processed by extracting individual frames using the OpenCV library. Each video was transformed into frames and resized to a uniform resolution of 300$\times$300 pixels to ensure consistency across the dataset. After frame extraction, the data was organized into feature-label pairs, with the features representing the video frames and the labels indicating the actions performed.

The dataset was divided into training and testing sets, with 80\% allocated for training and 20\% for evaluation. This division allowed the model to be trained on a substantial portion of the data while retaining a separate test set to assess performance on unseen samples. The CNN architecture included multiple convolutional layers followed by pooling layers to reduce dimensionality while preserving essential features. Dropout layers were integrated to mitigate overfitting, enhancing the model’s generalization capabilities.

The model was compiled using the categorical cross-entropy loss function, suitable for multi-class classification tasks, and the Adam optimizer was employed for effective training. Training was conducted over multiple epochs, with early stopping implemented to prevent overfitting by halting training when validation loss showed no improvement after several epochs.

Once trained, the model was evaluated on the test dataset, recording metrics such as accuracy and loss to gain insights into its performance. Graphical representations of accuracy and loss over time were created to visualize training progress, aiding in identifying potential overfitting issues. Additionally, the model was applied to real-time video prediction, showcasing its practical utility, particularly for the Surena Humanoid robot to detect given commands. Finally, the trained model was saved in a reusable format, ensuring quick loading for future predictions or refinements.

\subsection{LSTM Approach}

In the second approach, we processed a video dataset comprising various verbal commands for integration into a machine learning pipeline. The initial step involved transforming categorical labels into numerical representations to facilitate compatibility with machine learning models. This transformation was achieved through a custom function, which assigned unique integers to each verbal command (e.g., "begir," "benevis," "bia," etc.), creating a standardized numeric encoding necessary for supervised learning tasks.

The dataset was structured hierarchically, with each class of commands stored in its own directory. A Python script utilized to traverse the folder structure, collecting file paths for each video sample, which were stored in an array, while the corresponding class labels were stored in a separate array. This organization ensured the proper pairing of each video sample with its respective class label. The input and output arrays were then converted into NumPy arrays, optimizing them for numerical operations, crucial for efficient data handling in the machine learning pipeline.

The dataset was divided into three subsets—training, validation, and test data. The training set was designated for model optimization, while the validation set was used for tuning hyperparameters and preventing overfitting. The test set, remaining unseen during training, was reserved for final evaluation to assess the model’s generalization capability. A random state was set during the data splitting process to ensure reproducibility, a critical component of robust experimental design in machine learning.

Following data partitioning, the file paths and associated labels were prepared for input into the machine learning model, likely leveraging LSTM networks to analyze the visual data in the video files. The encoded numeric labels served as the target output for this supervised learning task, where the model was expected to learn the mapping between video input sequences and their respective command classes.

The preprocessed data was then fed into the model for training, anticipating that the deep learning architecture would extract meaningful features from the video frames. The lip movements captured in each video were processed as spatiotemporal data, enabling the model to recognize and classify spoken commands based on the visual input. The training process involved standard optimization techniques, such as stochastic gradient descent, with validation accuracy monitored to ensure robustness and avoid overfitting. Upon completion of training, the model's performance was evaluated using the test dataset, providing insights into its ability to generalize beyond the training data. Performance metrics— such as accuracy, precision, recall, and F1 score— were reported to gauge the effectiveness of the developed lip-reading recognition system.the sturctue of both approachs are illustrated in Fig. 2

\begin{figure*}[hbtp]  
    \centering
    \begin{minipage}{0.435\textwidth}
        \centering
        \includegraphics[width=\linewidth]{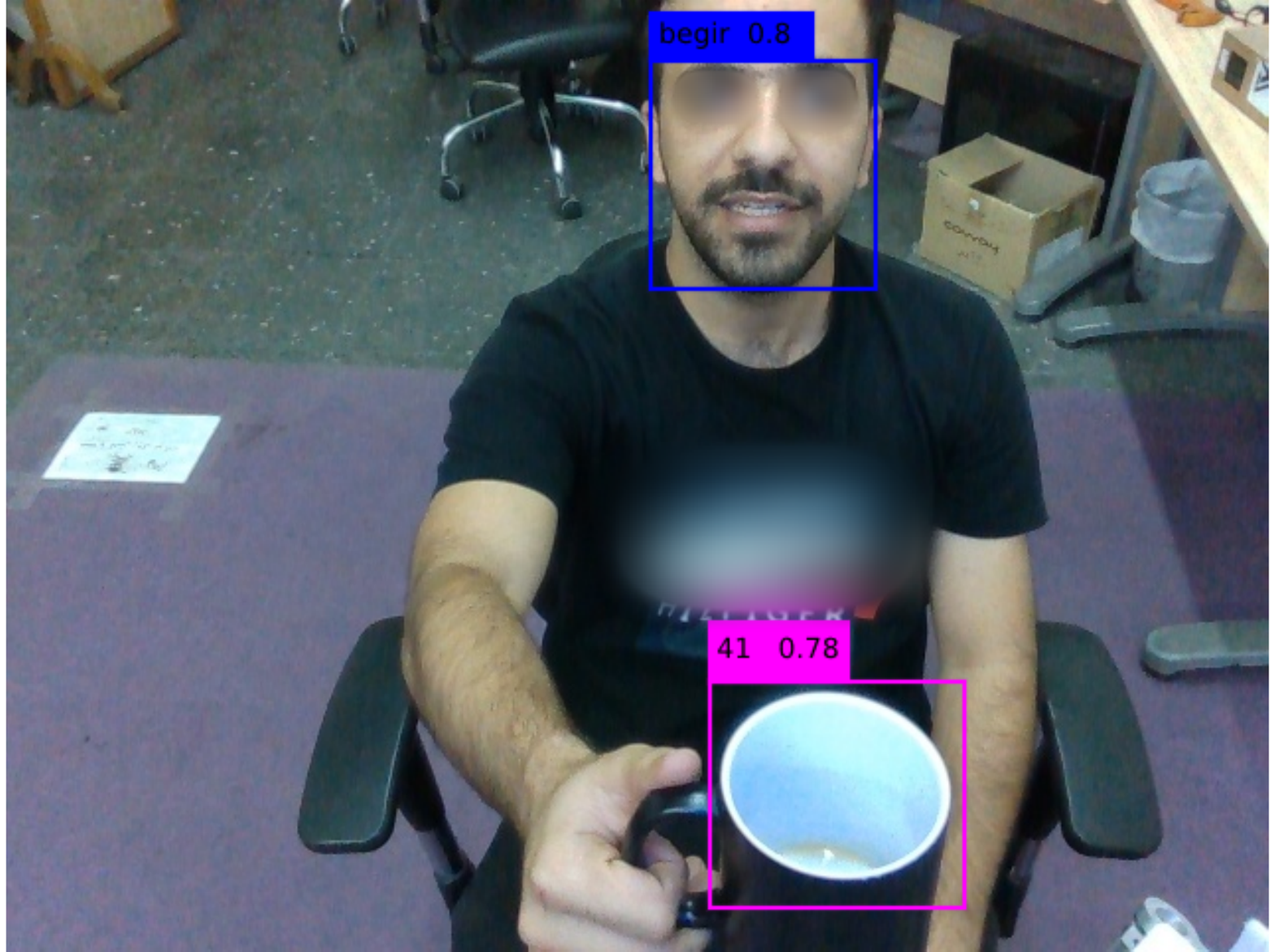}  
    \end{minipage}
    \hspace{0\textwidth}  
    \begin{minipage}{0.456\textwidth}
        \centering
        \includegraphics[width=\linewidth]{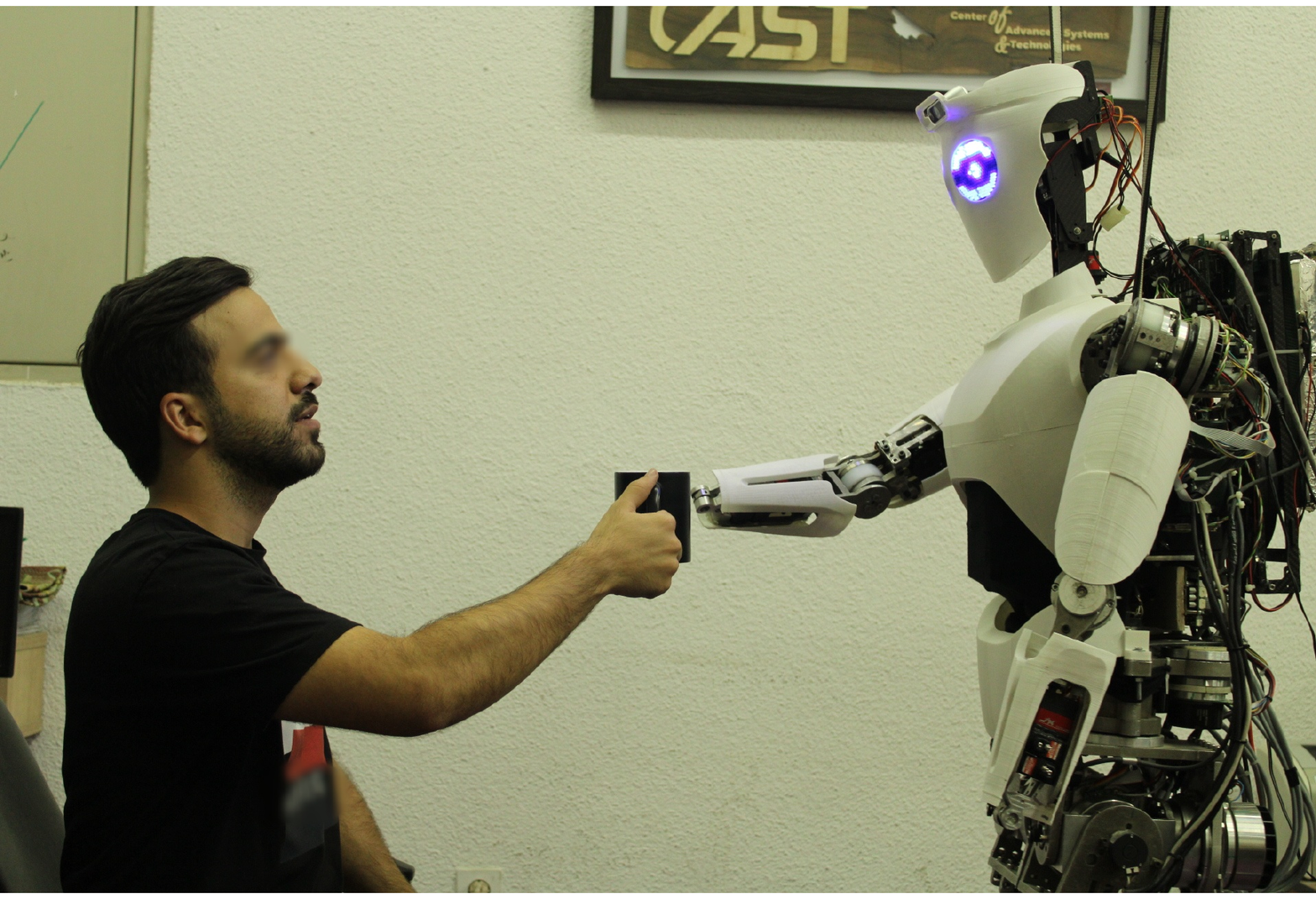}  
    \end{minipage}
    
    \vspace{0.008\textwidth}  
       
    \caption{The real-time implementation of LSTM Neural Network }
\end{figure*}

\section*{RESULTS}
The results for both the indirect and direct methods highlight the effectiveness of the approaches in detecting lip movements and classifying actions from video data.
\vspace{0.1cm}

\noindent A.Indirect Method Results

The indirect method, which utilizes facial landmarks and tracks lip movement, performed decent in recognizing speech patterns. The performance was evaluated across several CNN architectures, including VGG19, ResNet, and MobileNet. MobileNet emerged as the most effective model, achieving the highest accuracy and efficiency due to its lightweight architecture and depthwise separable convolutions.
MobileNet’s performance was particularly notable in its ability to process the 3D input data derived from facial landmarks. The accuracy of this model on the test set was consistently higher than other CNN models, with the highest accuracy of 52 \%  displayed in Table III . The model was also able to maintain a low error rate across various lighting conditions and minor head movements, thanks to the robust preprocessing steps that normalized the input data. These factors allowed MobileNet to handle variations in the dataset with minimal loss in accuracy.
In addition to its accuracy, MobileNet also excelled in speed, making it suitable for real-time lip-reading applications. The system processed the standardized one-second video portions efficiently, with minimal lag during live predictions.

\noindent B.direct Method Results

We employed a direct method focusing on action classification through Convolutional Neural Networks (CNNs) and Long Short-Term Memory networks (LSTMs) trained on raw video frames. The CNN model demonstrated impressive performance on the test dataset, achieving an accuracy of 75\%. Its architecture, comprising multiple convolutional and pooling layers, effectively captured the spatial hierarchies present in the input frames, contributing to its accurate classifications.

During the training phase, consistent improvements in both training and validation accuracy were observed, with minimal overfitting thanks to the implementation of dropout layers and early stopping mechanisms. Various evaluation metrics, including accuracy and loss, were closely monitored throughout the training process.

In contrast, the LSTM model surpassed the CNN in performance, achieving an accuracy of 89\% on the test set. This architecture, consisting of multiple time-distributed and dense layers, was more complex than the CNN, resulting in more precise predictions. Although some noise was noted during training—indicative of potential overfitting—this model ultimately converged effectively.

Visualizations of the models’ performance during training provided further insight into their learning capabilities. The loss and accuracy plots in Fig. 3 indicated steady convergence over time, highlighting the LSTM and CNN ability to learn more effectively from the data. A summary of the accuracy and loss metrics for both models is presented in Table III, reinforcing the overall strong performance achieved through our methodologies.

\noindent C.Real World Implementation

Notably,The LSTM model from the direct method has been successfully integrated into the Surena-V humanoid robot, enabling it to recognize and classify human order from video input. This real-world application highlights the robustness and practical utility of the method in facilitating advanced human-robot interactions. The system has achieved a prediction accuracy of 89\%, demonstrating that quick and accurate action recognition is critical for effective interaction between humans and robots.
Referring to Fig. 4, the robot successfully identified the word using the lip-reading model and detected the object through the YOLOv5 detector. Once both the word and object were recognized, the robot executed the appropriate, predefined response. In this case, the command was issued to the robot via the trained model, while other actions taken were beyond the scope of our study.

\begin{table}[htbp]
\caption{accuracy and loss of methods on last epoch}
\begin{center}
\begin{tabular}{|c|c|c|c|c|}
\hline
Method & Val-Acc & Train-Acc & Val-Loss & Train-Loss \\ \hline
Indirect   & 52       & 64      &   2.049       &  1.236   \\ \hline
Direct(CNN)     & 75       & 76      &   0.671     & 0.667  \\ \hline
Direct(LSTM)    & 95       & 93      &   0.6985    & 0.4702 \\ \hline
\end{tabular}
\label{tab1}
\end{center}
\end{table}

\section*{Conclusion}
The paper develops a Persian lip-reading system for the Surena-V humanoid robot using a custom dataset of participants pronouncing seven common words which was collected by 20 participants. It explores two methods: an indirect approach using facial landmark tracking and a direct approach utilizing CNN and LSTM networks to process raw video data. The innovation lies in the successful implementation of the LSTM model into the robot for real-time speech recognition. This work demonstrates how combining machine learning with robotics can enhance human-robot interaction, especially in settings where verbal communication is challenging.
The process of gathering data presented significant challenges, particularly in utilizing pre-trained models for lip and face detection, which often had constraints that hindered their effectiveness. For instance, while the best model could generate bounding boxes for faces, it struggled to perform accurately at certain distances. Additionally, the lip detector frequently returned null results during sudden head movements. Consequently, training the indirect method proved to be considerably more difficult than the direct approach.
For future work, the paper suggests enhancing the system by incorporating the optical flow estimation (Fedo) method to capture fine-grained lip movements across consecutive video frames. This approach would complement the current methods by providing more detailed motion dynamics, improving accuracy in recognizing subtle differences in lip movements. Additionally, expanding the dataset to include more words, phrases, and a wider variety of speakers is crucial for improving model generalization and robustness. Increasing the dataset's size, diversity, and complexity would help the system better handle variations in facial angles, lighting conditions, and background noise, further enhancing the Surena-V robot's real-time lip-reading performance.

\bibliography{references}

\begin{thebibliography}{99}
    \vspace{0.18cm}
    \bibitem{ref1}Dilip Kumar Margam, Rohith Aralikatti, Tanay Sharma, Abhinav Thanda, Pujitha A K, Sharad Roy, Shankar M Venkatesan. "Lip-Reading with 3D-2D-CNN BLSTM-HMM and word-CTC models" (2019). Available: \url{https://arxiv.org/pdf/1906.12170}
    
    \vspace{0.18cm}
    \bibitem{ref2} Suraj Paul, Dhanesh Lakhani, Divyanshu Aryan, Shudhashekhar Das, Rohit Varshney. "Lip Reading System for Speech-Impaired Individuals" (2024). Available: \url{https://www.researchgate.net/profile/Suraj-Paul4/publication/380297513_Lip_Reading_System_for_SpeechImpaired_Individuals/links/6634ca3f08aa54017ad5e61e/Lip-Reading-System-forSpeech-Impaired-Individuals.pdf}
    
    \vspace{0.18cm}
     \bibitem{ref3} D Parekh, A Gupta, S Chhatpar, A Yash, M Kulkarni. "Lip Reading Using Convolutional Auto Encoders as Feature Extractor" (2019). Available: \url{https://arxiv.org/pdf/1805.12371}
    
    \vspace{0.18cm}
    \bibitem{ref4} Sai Teja Krithik Putcha, Yelagandula Sai Venkata Rajam, K. Sugamya, Sushank Gopala. "Text Extraction and Translation Through Lip Reading using Deep Learning" (2024).
    
    \vspace{0.18cm}
    \bibitem{ref5} Marzieh Oghbaei, Arian Sabaghi, Kooshan Hashemifard, Mohammad Akbari. "Advances and challenges in deep lip reading" (2021). Available: \url{https://arxiv.org/pdf/2110.07879}
    
    \vspace{0.18cm}
   \bibitem{ref6} Pezhman Abdolahnezhad, Aghil Yousefi-Koma, Amirhosein Vedadi, Kasra Sinaei, Behnam Maleki, Milad Shafiee. "Online Bipedal Locomotion Adaptation for Stepping on Obstacles Using a Novel Foot Sensor" (2022). Available: \url{https://arxiv.org/pdf/2212.13416}
    
    \vspace{0.18cm}
     \bibitem{ref7} Rabbia Mahum, Faisal Shafique Butt, Kashif Ayyub, Seema Islam, Marriam Nawaz, Daud Abdullah. "A review on humanoid robots" (2017). Available: \url{https://www.academia.edu/download/71902079/15_202017-4-2-pp.83-90.pdf}
    
    \vspace{0.18cm}
    \bibitem{ref8} Javad Peymanfard, Ali Lashini, Samin Heydarian, Hossein Zeinali, Nasser Mozayani. "Word-level Persian Lipreading Dataset" (2023). Available: \url{https://arxiv.org/pdf/2304.04068}
    
    \vspace{0.18cm}
     \bibitem{ref9} Mahsa Hedayatipour, Yaser Shekofte, Mohsen Ebrahimi Moghaddam. "Persian Audi Visual Database of CV syllables" (2021). Available: \url{https://ieeexplore.ieee.org/abstract/document/9544268}
    
    \vspace{0.18cm}
    \bibitem{ref10} Talya TUMER SIVRI, Ali BERKOL, Hamit ERDEM. "Lip Reading Using Various Deep Learning Models with Visual Turkish Data" (2024). Available: \url{https://dergipark.org.tr/en/download/article-file/2904250}
    
    \vspace{0.18cm}
   \bibitem{ref11} Aripin, Abas Setiawan. "Indonesian Lip-Reading Detection and Recognition Based on Lip Shape Using Face Mesh and Long-Term Recurrent Convolutional Network" (2024). Available: \url{https://scholar.google.com/scholar?output=instlink&q=info:spDJeCnR1igJ:scholar.google.com/&hl=en&as_sdt=0,5&scillfp=2163467829437430261&oi=lle}
    
    \vspace{0.18cm}
    \bibitem{ref12} Muhammad Faisal, Sanaullah Manzoor. "Deep Learning for Lip Reading using Audio-Visual Information for Urdu Language" (2018). Available: \url{https://arxiv.org/pdf/1802.05521}
    
    \vspace{0.18cm}
    \bibitem{ref13} Waleed Dweik, Sundus Altorman, Safa Ashour. "Read my lips: Artificial intelligence word-level arabic lipreading system" (2022). Available: \url{https://www.sciencedirect.com/science/article/pii/S1110866522000433}

    \vspace{0.18cm}
    \bibitem{ref14} Newcombe, R.A., Izadi, S., et al., "KinectFusion: Real-time 3D Reconstruction and Interaction Using a Moving Depth Camera," Proceedings of the 24th ACM Symposium on User Interface Software and Technology, 2011.

    \vspace{0.18cm}
    \bibitem{ref15}Shaha, Manali, and Meenakshi Pawar. "Transfer learning for image classification." 2018 second international conference on electronics, communication and aerospace technology (ICECA). IEEE, 2018.
\end{thebibliography}

\end{document}